\documentclass[letterpaper]{article}
\usepackage{aaai}
\usepackage{times}
\usepackage{helvet}
\usepackage{courier}
\usepackage{url}
\usepackage{amsmath}
\usepackage{amsbsy}
\usepackage{amssymb}
\usepackage{amsopn}
\usepackage{stmaryrd}
\usepackage{graphicx}
\usepackage{multirow}
\usepackage{rotating}
\usepackage{algorithm,algpseudocode}
\newcommand{\PP}{
\mathrm{P}
}
\begin{document}
%
\title{A Dataset for StarCraft AI \& an Example of Armies Clustering}
\author{
Gabriel Synnaeve and Pierre Bessi\`{e}re \\
Coll\`{e}ge de France, CNRS, Grenoble University, LIG\\
\texttt{gabriel.synnaeve@gmail.com} and \texttt{pierre.bessiere@imag.fr}
}
\maketitle
\begin{abstract}
This paper advocates the exploration of the full state of recorded real-time strategy (RTS) games, by human or robotic players, to discover how to reason about tactics and strategy. We present a dataset of StarCraft\footnote{StarCraft and its expansion StarCraft: Brood War are trademarks of Blizzard Entertainment$^{\mathrm{TM}}$} games encompassing the most of the games' state (not only player's orders). We explain one of the possible usages of this dataset 
by clustering 
armies on their compositions. This reduction of armies compositions to mixtures of Gaussian allow for strategic reasoning at the level of the components. We evaluated this clustering method by predicting the outcomes of battles based on armies compositions' mixtures components.
\end{abstract}

\section{Introduction}
Real-time strategy (RTS) games AI is not yet at a level high enough to compete with trained/skilled human players. Particularly, adaptation to different strategies (of which army composition) and to tactics (army moves) are strong indicators of human-played games \cite{HagelbackCIG10}. So, while micro-management (low level units control) has known tremendous improvements in recent years, the broadest high-level strategic reasoning is not yet an exemplary feature neither of commercial games nor of StarCraft AI competitions' entries. At best, StarCraft bots have an estimation of the available technology of their opponents and use rules encoding players' knowledge to adapt their strategy. We believe that better strategic reasoning is a matter of abstracting and combining the low level states at an expressive higher level of reasoning. Our approach will be to learn unsupervised representations of low-level features.

We worked on StarCraft: Brood War, which is a canonical RTS game, as Chess is to board games. It had been around since 1998, has sold 10 millions licenses and was the best competitive RTS for more than a decade. There are 3 factions (Protoss, Terran and Zerg) that are totally different in terms of units, build trees / tech trees (directed acyclic graphs of the buildings and technologies) and thus gameplay styles. StarCraft and most RTS games provide a tool to record game logs into \textit{replays} that can be re-simulated by the game engine. That is this trace mechanism that we used to download and simulate games of professional gamers and highly skilled international competitors.

This paper is separated in two parts. The first part explains what is in the dataset of StarCraft games that we put together. The second part showcases army composition reduction to a mixture of Gaussian distributions, and give some evaluation of this clustering.

\section{Related Work}
There are several ways to produce strategic abstractions: from using high-level gamers' vocabulary, and the game rules (build/tech trees), to salient low-level (shallow) features. Other ways include combining low-level and higher-level strategic representation and/or interdependencies between states and sequences. 

Case-based reasoning (CBR) approaches often use extensions of build trees as state lattices (and sets of tactics for each state) as for \cite{LTW,GA} in Wargus. \citeauthor{OntanonCBR} (\citeyear{OntanonCBR}) base their real-time case-based planning (CBP) system on a plan dependency graph which is learned from human demonstration in Wargus. In \cite{PlanRetrieval}, they use ``situation assessment for plan retrieval'' from annotated replays, which recognizes distance to behaviors (a goal and a plan), and selected only the low-level features with the higher information gain. 
\citeauthor{HsiehS08} (\citeyear{HsiehS08}) based their work on \cite{LTW} and used StarCraft replays to construct states and building sequences. Strategies are choices of building construction order in their model. 

\citeauthor{schadd2007opponent} (\citeyear{schadd2007opponent}) describe opponent modeling through hierarchically structured models of the opponent behavior and they applied their work to the Spring RTS game (Total Annihilation open source clone). 
\citeauthor{UCT} (\citeyear{UCT}) applied upper confidence bounds on trees (UCT: a Monte-Carlo planning algorithm) to tactical assault planning in Wargus, their tactical abstraction combines units hit points and locations. 
In \cite{SYNNAEVE:StratPred}, they 
predict the build trees 
of the opponent a few buildings before they are built. Another approach is to use the gamers' vocabulary of strategies (and openings) to abstract even more what strategies represent (a set of states, of sequences and of intentions) as in \cite{weberStrat,SYNNAEVE:OpeningPred}. 
\citeauthor{HMMstrat_RTS_AIIDE11} (\citeyear{HMMstrat_RTS_AIIDE11}) used an hidden Markov model (HMM) whose states are extracted from (unsupervised) maximum likelihood on a StarCraft dataset. The HMM parameters are learned from unit counts (both buildings and military units) every 30 seconds and ``strategies'' are the most frequent sequences of the HMM states according to observations.

Few models have incorporated army compositions in their strategy abstractions, except sparsely as an aggregate or boolean existence of unit types. Most strategy abstractions are based on build trees (or tech trees), although a given set of buildings can produce different armies. What we will present here is complementary to these strategic abstractions and should help the military situation assessment.

\section{Dataset}
We downloaded more than 8000 replays to keep 7649 uncorrupted, 1 vs. 1 replays from professional gamers leagues and international tournaments of StarCraft, from specialized websites\footnote{\url{http://www.teamliquid.net}}\footnote{\url{http://www.gosugamers.net}}\footnote{\url{http://www.iccup.com}}. We then ran them using Brood War API\footnote{BWAPI \url{http://code.google.com/p/bwapi/}} and dumped: units' positions, regions' positions, pathfinding distance between regions, resources (every 25 frames), all players' orders, vision events (when units are seen) and attacks (types, positions, outcomes). Basically, we recorded every BWAPI event, plus interesting states and attacks. The dataset is freely available\footnote{\url{http://emotion.inrialpes.fr/people/synnaeve/TLGGICCUP_gosu_data.7z}}
, the source code and a documentation are also provided\footnote{\url{http://snippyhollow.github.com/bwrepdump/}}


\subsection{Regions}
\citeauthor{Forbus2002} (\citeyear{Forbus2002}) have shown the importance of qualitative spatial reasoning, and it would be too space-consuming to dump the ground distance of every position to any other position. For these reasons, we discretized StarCraft maps in two types of regions:
\begin{itemize}
    \item Brood War Terrain Analyzer\footnote{BWTA \url{http://code.google.com/p/bwta/}} produced regions from a pruned Voronoi diagram on walkable terrain \cite{Perkins2010}. Chokes are the boundaries of such regions.
    \item As battles often happens at chokes, we also produced choke-dependent regions (CDR), which are created by doing an additional (distance limited) Voronoi tessellation spawned at chokes. This regions set is $$CDR = (regions \setminus chokes) \cup chokes$$
\end{itemize}

\subsection{Attacks}
We trigger an attack tracking heuristic when one unit dies and there are at least two military units around. We then update this attack until it ends, recording every unit which took part in the fight. We log the position, participating units and fallen units for each player, the attack type and of course the attacker and the defender. Algorithm~\ref{alg:attackheuristic} 
shows how we detect attacks.

We annotated attacks by four types (but researchers can also produce their own annotations given the state available):
\begin{itemize}
    \item \textit{ground} attacks, which may use all types of units (and so form the large majority of attacks).
    \item \textit{air} raids, air attacks, which can use only flying units.
    \item \textit{invisible} (ground) attacks, which can use only a few specific units in each race (Protoss Dark Templars, Terran Ghosts, Zerg Lurkers).
    \item \textit{drop} attacks, which need a transport unit (Protoss Shuttle, Terran Dropship, Zerg Overlord with upgrade).
\end{itemize}

\begin{algorithm}
\caption{Simplified attack tracking heuristic for extraction from games. The heuristics to determine the attack type and the attack radius and position are not described here. They look at the proportions of units types, which units are firing and the last actions of the players.}
\label{alg:attackheuristic}
\begin{algorithmic}
\State $\mathrm{\textbf{list}}\ tracked\_attacks$
\Function{unit\_death\_event}{$unit$}
    \State $tmp \leftarrow tracked\_attacks.which\_contains(unit)$
    \If{$tmp \neq \emptyset$}
        \State $tmp.update(unit)$ \Comment{$\Leftrightarrow update(tmp, unit)$}
    \Else
        \State $tracked\_attacks.push(attack(unit))$
    \EndIf
\EndFunction

\Function{attack}{$unit$} \Comment{new attack constructor}
    \State $ $ \Comment{$self \Leftrightarrow this$}
    \State $self.convex\_hull \leftarrow default\_hull(unit)$ 
    \State $self.type \leftarrow attack\_type(update(self, unit))$
    \State \Return $self$
\EndFunction

\Function{update}{$attack, unit$}
    \State $attack.update\_hull(unit)$ \Comment{takes units ranges into account}
    \State $c \leftarrow get\_context(attack.convex\_hull)$
    \State $self.units\_involved.update(c)$ 
    \State $self.tick \leftarrow default\_timeout()$
    \State \Return $c$
\EndFunction

\Function{tick\_update}{}
    \State $self.tick \leftarrow self.tick - 1$
    \If{$self.tick < 0$}
        \State $self.destruct()$
    \EndIf
\EndFunction
\end{algorithmic}
\end{algorithm}

\subsection{Information in the dataset}
Table~\ref{tab:dataset} shows some metrics about the dataset. Note that the numbers of attacks for a given race have to be divided by (approximatively) two in a given \textit{non-mirror} match-up. So, there are 7072 Protoss attacks in PvP and there are not 70,089 attacks by Protoss in PvT but about half that.
\begin{table*}[ht]
\begin{center}
\begin{tabular}{|l|c|c|c|c|c|c|}
\hline
match-up & PvP & PvT & PvZ & TvT & TvZ & ZvZ \\
\hline
number of games & 445 & 2408 & 2027 & 461 & 2107 & 199 \\
number of attacks & 7072 & 70089 & 40121 & 16446 & 42175 & 2162 \\
mean attacks/game & 15.89 & 29.11 & 19.79 & 35.67 & 20.02 & 10.86 \\
mean time (frames) / game & 32342 & 37772 & 39137 & 37717 & 35740 & 23898 \\
mean time (minutes) / game & 22.46 & 26.23 & 27.18 & 26.19 & 24.82 & 16.60 \\
actions issued (game engine) / game & 24584 & 33209 & 31344 & 26998 & 29869 & 21868 \\
mean regions / game & 19.59 & 19.88 & 19.69 & 19.83 & 20.21 & 19.31 \\
mean CDR / game & 41.58 & 41.61 & 41.57 & 41.44 & 42.10 & 40.70 \\
mean ground distance\footnote{for regions connected by ground, pathfinding aware, in pixels} region $\leftrightarrow$ region & 2569 & 2608 & 2607 & 2629 & 2604 & 2596 \\ 
mean ground distance\footnote{for choke-dependent regions connected by ground, pathfinding aware, in pixels} CDR $\leftrightarrow$ CDR & 2397 & 2405 & 2411 & 2443 & 2396 & 2401 \\ 
\hline
\end{tabular}
\end{center}
\caption{Detailed numbers about our dataset. XvY means race X vs race Y matches and is an abbreviation of the match-up: PvP stands for Protoss versus Protoss.}
\label{tab:dataset}
\end{table*}

By running the recorded games (replay) through StarCraft, we were able to recreate the full state of the game. Time is always expressed in game frames (24 frames per second). We recorded three types of files:
\begin{itemize}
    \item general data (\texttt{*.rgd} files): records the players' names, the map's name, and all information about events like \textit{creation} (along with \textit{morph}), \textit{destruction}, \textit{discovery} (for one player), \textit{change of ownership} (special spell/ability), for each units. It also shows attack events (detected by a heuristic, see below) and dumps the current economical situation every 25 frames: mineral, gas, supply (count and total: maxsupply).
    \item order data (\texttt{*.rod} files): records all the orders which are given to the units (individually) like \textit{move}, \textit{harvest}, \textit{attack unit}, the orders positions and their issue time.
    \item location data (\texttt{*.rld} files): records positions of mobile units every 100 frames, and their position in regions and choke-dependent regions if they changed since last measurement. It also stores ground distances (pathfinding-wise) matrices between regions and choke-dependent regions in the header.
\end{itemize}
From this data, one can recreate most of the state of the game: the map key characteristics (or load the map separately), the economy of all players, their \textit{tech} (all researches and upgrades), all the buildings and units, along with their orders and their positions.

\section{Armies composition}

We will consider units engaged in these attacks as armies and will seek a compact description of armies compositions.

\subsection{Armies clustering}

The idea behind armies clustering is to give one ``composition'' label for each army depending on its composing ratio of the different unit types. Giving a ``hard'' (unique) label for each army does not work well because armies contain different components of unit types combinations. For instance, a Protoss army can have only a ``Zealots+Dragoons'' component, but it will often just be one of the components (sometimes the backbone) of the army composition, augmented for instance with ``High Templars+Archons''. 

Because a hard clustering is not an optimal solution, we used a Gaussian mixture model (GMM), which assumes that an army is a mixture (i.e. weighted sum) of several (Gaussian) components. We present the model in the Bayesian programming framework \cite{Diard03}: we first describe the variables, the decomposition (independence assumptions) and the forms of the distribution. Then, we explain how we identified (learned) the parameters and lay out the question that we will ask this model in the following parts.

\subsubsection{Variables}
    \begin{itemize}
        \item $C \in \llbracket c_1\dots c_K \rrbracket$, our army clusters/components ($C$). There are $K$ units clusters and $K$ depends on the race (the mixture components are not the same for Protoss/Terran/Zerg).

        \item $U \in ([0,1]\dots[0,1])$ (length $N$), our $N$ dimensional unit types ($U$) proportions, i.e. $U \in [0,1]^N$. $N$ is dependent on the race and is the total number of unit types. For instance, an army with equal numbers of $Zealots$ and $Dragoons$ (and nothing else) is represented as $\{U_{Zealot}=0.5, U_{Dragoon}=0.5, \forall ut \neq Zealot|Dragoon\ U_{ut}=0.0\}$, i.e. $U=(0.5,0.5,0,\dots,0)$ if $Zealots$ and $Dragoons$ are the first two components of the $U$ vector. So $\sum_i U_i = 1$ whatever the composition of the army.
    \end{itemize}

\subsubsection{Decomposition:}

For the $M$ battles, the armies compositions are independent across battles, and the unit types proportions vector (army composition) is generated by a mixture of Gaussian components and thus $U_i$ depends on $C_i$.
$$\PP(U_{1 \dots M}, C_{1 \dots M}) = \prod_{i=1}^M \PP(U_i|C_i)\PP(C_i)$$

\subsubsection{Forms}
    \begin{itemize}
        \item $\PP(U_i|C_i)$ mixture of Gaussian distributions:
        \begin{eqnarray*}
            \PP(U_i|C_i=c) = \mathcal{N}(\mu_{c}, \sigma_{c}^2)\\
        \end{eqnarray*}
        \item $\PP(C_i) = Categorical(K, p_{C})$: 
        \begin{eqnarray*}
            \begin{cases} \PP(C_i=c_k) = p_{k}\\
            \sum_{k=1}^K p_k = 1 \end{cases}
        \end{eqnarray*}
    \end{itemize}

\subsubsection{Identification (learning):}

We learned the Gaussian mixture models (GMM) parameters with the expectation-maximization (EM) algorithm on 5 to 15 mixtures with spherical, tied, diagonal and full co-variance matrices, using scikit-learn \cite{scikit-learn}. We kept the best scoring models (by varying the number of mixtures) according to the Bayesian information criterion (BIC) \cite{schwarz1978}. 

    Let $\theta= (\mu_{1:K},\sigma^2_{1:K})$, being respectively the $K$ different $N$-dimensional means ($\mu_{1:K}$) and the variances ($\sigma^2_{1:K}$) of the normal distributions. Initialize $\theta$ randomly, and let $$L(\theta;U) = \PP(U|\theta) = \prod_{i=1}^M \sum_{k=1}^K\PP(U_i|\theta,C_i=c_k)\PP(C_i=c_k)$$
Iterate until convergence (of $\theta$):
    \begin{enumerate}
        \item E-step: $Q(\theta|\theta^{(t)})= \mathrm{E}[\log L(\theta;u,C)]$ $$= \mathrm{E}\big[\log \prod_{i=1}^M \sum_{k=1}^K\PP(u_i|C_i=c_k,\theta)\PP(C_i=c_k)\big]$$
        \item M-step: $\theta^{(t+1)}=\mathrm{argmax}_{\theta} Q(\theta|\theta^{(t)})$
    \end{enumerate}

\subsubsection{Question:} 

For the $i$th battle (one army with units $u$):
$$\PP(C_i | U_i=u) = \PP(C_i) \PP(U_i=u|C_i)$$

\subsection{Counter compositions}

In a battle, there are two armies (one for each players), we can thus apply this clustering to both the armies. If we have $K$ clusters and $N$ unit types, the opponent has $K'$ clusters and $N'$ unit types. We introduce $EU$ and $EC$, respectively with the same semantics as $U$ and $C$ but for the enemy. In a given battle, we observe $u$ and $eu$, respectively our army composition and the enemy's army composition. We can ask $\PP(C|U=u)$ and $\PP(EC|EU=eu)$. 

As StarCraft unit types have strengths and weaknesses against other types, we can learn which clusters should beat other clusters (at equivalent investment) as a probability table. We use Laplace's law of succession (``add-one smoothing'') by counting and weighting according to battles results ($c>ec$ means ``$c$ beats $ec$'', i.e. we won against the enemy):
$$\PP(C=c | EC=ec) = \frac{1 + \PP(c)\PP(ec) \mathrm{count_{battles}}(c > ec)}{K + \PP(ec)\mathrm{count_{battles\ with}}(ec)}$$

\subsection{Results}

We used the dataset presented in this paper to learn all the parameters and perform the benchmarks (by setting 100 test matches aside and learning on the remaining of the dataset). First, we analyze the posteriors of clustering only one army and then we evaluated the clustering as a mean to predict outcomes of battles.


\subsubsection{Posterior analysis:}
\begin{figure*}[ht]
\centerline{\includegraphics[width=2\columnwidth]{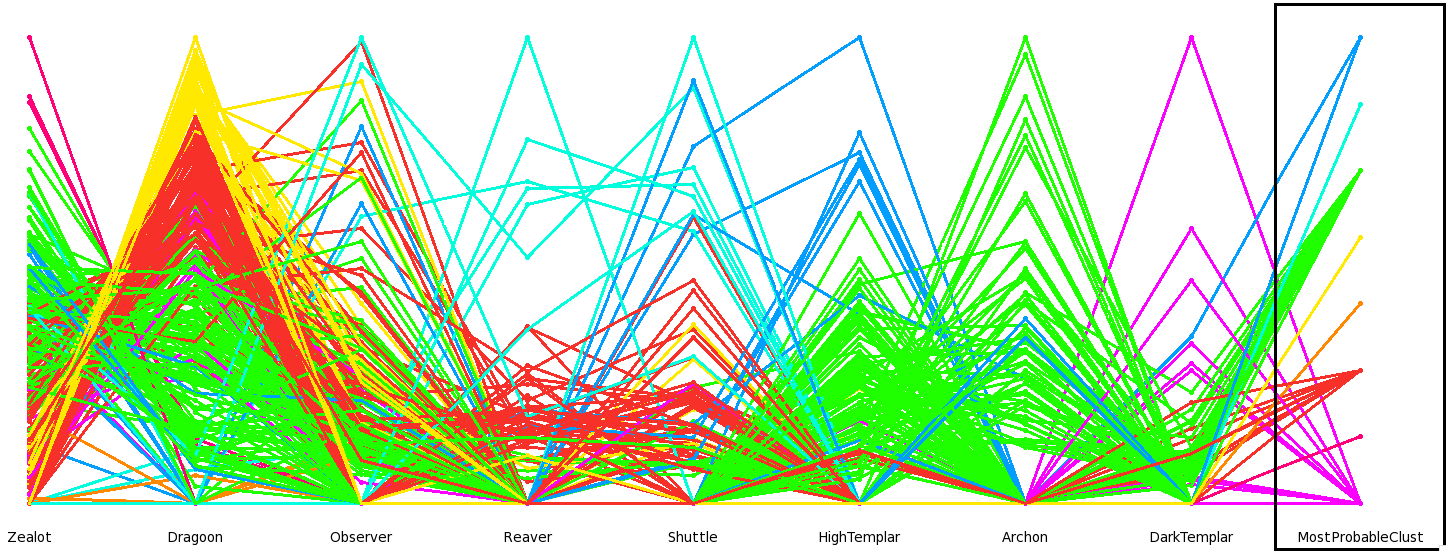}}
\caption{Parallel plot of a small dataset of Protoss (vs Protoss, i.e. in the PvP match-up) army clusters on most important unit types (for the match-up). Each normalized vertical axis represents the percentage of the units of the given unit type in the army composition (we didn't remove outliers, so most top vertices (tip) represent 100\%), except for the rightmost (framed) which links to the most probable GMM component. Note that several traces can (and do) go through the same edge.}
\label{fig:parallelplot}
\vspace{-0.3cm}
\end{figure*}

Figure~\ref{fig:parallelplot} shows a parallel plot of \textit{army compositions}. We removed the less frequent unit types to keep only the 8 most important unit types of the PvP match-up, and we display a 8 dimensional representation of the army composition, each vertical axis represents one dimension. Each line (trajectory in this 8 dimensional space) represents an army composition (engaged in a battle) and gives the percentage 
of each of the unit types. These lines (armies) are colored with their most probable mixture component, which are shown in the rightmost axis. We have 8 clusters (Gaussian mixtures components): this is not related to the 8 unit types used as the number of mixtures was chosen by BIC score. Expert StarCraft players will directly recognize the clusters of typical armies, here are some of them:
\begin{itemize}
    \item Light blue corresponds to the ``Reaver Drop'' tactical squads, which aims are to transport (with the flying Shuttle) the slow Reaver (zone damage artillery) inside the opponent's base to cause massive economical damages.
    \item Red corresponds to the ``Nony'' typical army that is played in PvP (lots of Dragoons, supported by Reaver and Shuttle).
    \item Green corresponds to a High Templar and Archon-heavy army: the gas invested in such high tech units makes it that there are less Dragoons, completed by more Zealots (which cost no gas).
    \item Purple corresponds to Dark Templar (``sneaky'', as Dark Templars are invisible) special tactics (and opening).
\end{itemize}

\begin{figure}[h]
\centerline{\includegraphics[width=1\columnwidth]{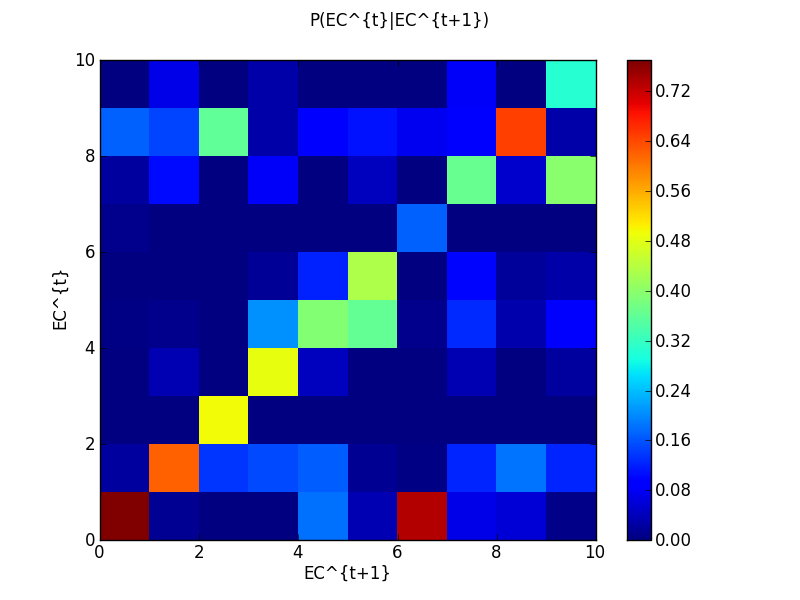}}
\caption{Dynamics of clusters: $\PP(EC^t|EC^{t+1})$ for Zerg, with $\Delta t = 2$ minutes}
\label{ecknowingecnext}
\vspace{-0.3cm}
\end{figure}

Figure~\ref{ecknowingecnext} showcases the dynamics of clusters components: $\PP(EC^t|EC^{t+1}$, for Zerg (vs Protoss) for $\Delta t$ of 2 minutes. The diagonal components correspond to those which do not change between $t$ and $t+1$ ($\Leftrightarrow t+2$minutes), and so it is normal that they are very high. The other components show the shift between clusters. For instance, the first line seventh column (in (0,6)) square shows a brutal transition from the first component (0) to the seventh (6). This may be the production of Mutalisks\footnote{Mutalisks are flying units which require to unlock several technologies and thus for which player save up for the production while opening their tech tree.} from a previously very low-tech army (Zerglings).

\subsubsection{A soft rock-paper-scissors:}

We then used the learned $\PP(C|EC)$ table to estimate the outcome of the battle. For that, we used battles with limited \textit{disparities} (the maximum strength ratio of one army over the other) of 1.1 to 1.5. Note that the army which has the superior forces numbers has \textbf{more than a linear advantage} over their opponent (because of focus firing\footnote{Efficiently micro-managed, an army 1.5 times superior to their opponents can keep much more than one third of the units alive.}), so a disparity of 1.5 is very high. For information, there is an average of 5 battles per game at a 1.3 disparity threshold, and the numbers of battles (used) per game increase with the disparity threshold.

We also made up a baseline heuristic, which uses the sum of the values of the units to decide which side should win. If we note $v(unit)$ the value of a unit, the heuristic computes $\sum_{unit} v(unit)$ for each army and predicts that the winner is the one with the biggest score. For the value of a unit we used:  $$v(unit) = minerals\_value + \frac{4}{3}gas\_value + 50supply$$ Of course, we recall that a random predictor would predict the result of the battle correctly $50\%$ of the time.

A summary of the main metrics is shown in Table~\ref{tab:openingsresults}, the first line can be read as: for a forces disparity of 1.1, for Protoss vs Protoss (first column),
\begin{itemize}
    \item considering only military units
\begin{itemize}
    \item the heuristic predicts the outcome of the battle correctly 63\% of the time.
    \item the probability of a clusters mixture to win against another ($\PP(C|EC)$), without taking the forces sizes into account, predicts the outcome correctly 54\% of the time.
    \item the probability of a clusters mixture to win against another, taking also the forces sizes into account ($\PP(C|EC)\times \sum_{unit}v(unit)$), predicts the outcome correctly 61\% of the time.
\end{itemize}
    \item considering only all units involved in the battle (military units, plus static defenses and workers): same as above.
\end{itemize}
Results are given for all match-up (columns) and different forces disparities (lines). The last column sums up the means on all match-ups, with the whole army (military units plus static defenses and workers involved), for the three metrics.

Also, without explicitly labeling clusters, one can apply thresholding to special units (Observers, Arbiters, Defilers...) to generate more specific clusters: we did not put these results here (they include too much expertize/tuning) but they sometimes drastically increase prediction scores, as one Observer can change the course of a battle.

\setlength{\tabcolsep}{6pt}
\begin{table*}[ht]
\begin{center}
\begin{small}
\begin{tabular}{|c|c|cc|cc|cc|cc|cc|cc|c|}
\hline
\begin{footnotesize}forces\end{footnotesize} & scores & \multicolumn{2}{|c|}{PvP} & \multicolumn{2}{|c|}{PvT} & \multicolumn{2}{|c|}{PvZ} & \multicolumn{2}{|c|}{TvT} & \multicolumn{2}{|c|}{TvZ} & \multicolumn{2}{|c|}{ZvZ} & mean \\
\begin{footnotesize}disparity\end{footnotesize} & in \% & m & ws & m & ws & m & ws & m & ws & m & ws& m & ws & ws\\
\hline

& heuristic & \textbf{63} & 63 & 58 & 58 & 58 & 58 & \textbf{65} & \textbf{65} & 70 & 70 & 56 & 56 & 61.7 \\
1.1     & \textbf{just prob.} & 54 & 58 & 68 & \textit{72} & 60 & 61 & 55 & 56 & 69 & 69 & 62 & 63 & 63.2 \\ 
    & prob$\times$heuristic & 61 & \textbf{63} & \textbf{69} & \textbf{72} & \textbf{59} & \textbf{61} & 62 & 64 & \textbf{70} & \textbf{73} & \textbf{66} & \textbf{69} & 67.0 \\
\hline
& heuristic & \textbf{73} & 73 & 66 & 66 & \textbf{69} & \textbf{69} & \textbf{75} & 72 & 72 & \textbf{72} & 70 & 70 & 70.3 \\
1.3     & \textbf{just prob.} & 56 & 57 & 65 & \textit{66} & 54 & 55 & 56 & 57 & 62 & 61 & 63 & 61 & 59.5 \\
    & prob$\times$heuristic & 72 & \textbf{73} & \textbf{70} & \textbf{70} & 66 & 66 & 71 & \textbf{72} & \textbf{72} & 70 & \textbf{75} & \textbf{75} & 71.0 \\
\hline
& heuristic & 75 & 75 & 73 & 73 & \textbf{75} & \textbf{75} & 78 & \textbf{80} & \textbf{76} & 76 & 75 & 75 & 75.7 \\
1.5     & \textbf{just prob.} & 52 & 55 & 61 & 61 & 54 & 54 & 55 & 56 & 61 & \textit{63} & 56 & 60 & 58.2 \\
    & prob$\times$heuristic & \textbf{75} & \textbf{76} & \textbf{74} & \textbf{75} & 72 & 72 & \textbf{78} & 78 & 73 & \textbf{76} & \textbf{77} & \textbf{80} & 76.2 \\
\hline
\end{tabular}
\end{small}
\caption{Winner prediction scores (in \%) for the three main metrics. For the left columns (``m''), we considered only military units. For the right columns (``ws'') we also considered static defense and workers. The ``heuristic'' metric is a baseline heuristic for battle winner prediction for comparison using army values, while ``just prob.'' only considers $\PP(C|EC)$ to predict the winner, and ``prob$\times$heuristic'' balances the heuristic's predictions with $\sum_{C,EC}\PP(C|EC)\PP(EC)$.}
\label{tab:openingsresults}
\end{center}
\vspace{-0.5cm}
\end{table*}

We can see that predicting battle outcomes (even with a high disparity) with ``just probabilities'' of $\PP(C|EC)$ (without taking the forces into account) gives relevant results as they are always above random predictions. Note that this is a very high level (abstract) view of a battle, we do not consider tactical positions, nor players' attention, actions, etc. Also, it is better (in average) to consider the heuristic with the composition of the army (``prob$\times$heuristic'') than to consider the heuristic alone, even for high forces disparity. Our heuristic augmented with the clustering seem to be the best indicator for battle situation assessment. These prediction results with ``just prob.'', or the fact that heuristic with $\PP(C|EC)$ tops the heuristic alone, are a proof that the assimilation of armies compositions as Gaussian mixtures of cluster works.

Secondly, and perhaps more importantly, we can view the difference between ``just prob.'' results and random guessing (50\%) as the military \textbf{efficiency improvement} that we can (at least) expect from having the right army composition. 
Indeed, for small forces disparities (up to 1.1 for instance), the prediction based only on army composition (``just prob.'': 63.2\%) is better than the prediction with the baseline heuristic (61.7\%). It means that \textit{we can expect to win 63.2\% of the time (instead of 50\%) with an (almost) equal investment if we have the right composition}. 
Also, when we predict 58.5\% of the time the accurate result of a battle with disparity up to 1.5 from ``just prob.'', this success in prediction is independent of the sizes of the armies. What we predicted is that the player with the better army composition won (and not necessarily the one with more or more expensive units).

\section{Conclusion}
We delivered a rich StarCraft dataset which enables the study of tactical and strategic elements of RTS gameplay. Our (successful) previous works on this dataset include learning a tactical model of where and how to attack (both for prediction and decision-making), and the analysis of units movements. We provided the source code of the extracting program (using BWAPI), which can be run on other replays. We proposed and validated an encoding of armies composition which enables efficient situation assessment and strategy adaptation. We believe it can benefit all the current StarCraft AI approaches. Moreover, the probabilistic nature of the model make it deal natively with incomplete information about the opponent's army.

\bibliographystyle{aaai}
\bibliography{workshop}

\end{document}